\documentclass{article}

\usepackage[final]{neurips_2024}

\usepackage{amsmath,amsfonts,bm}

\def\eqref#1{equation~\ref{#1}}

\def\1{\bm{1}}

\DeclareMathAlphabet{\mathsfit}{\encodingdefault}{\sfdefault}{m}{sl}
\SetMathAlphabet{\mathsfit}{bold}{\encodingdefault}{\sfdefault}{bx}{n}

\usepackage{booktabs}       
\usepackage{multirow}

\newcommand{\mr}[2]{\multirow{#1}{*}{#2}}
\newcommand{\mc}[3]{\multicolumn{#1}{#2}{#3}}
\usepackage{amsfonts}       
\usepackage{nicefrac}       
\usepackage{microtype}      
\usepackage{array}
\newcolumntype{P}[1]{>{\centering\arraybackslash}p{#1}}
\usepackage{url}            

\usepackage{amsmath,amsthm} 
\usepackage{xspace}
\usepackage{enumerate}
\usepackage{enumitem}
\usepackage{stmaryrd}
\usepackage{caption}
\usepackage{subcaption}
\usepackage[pdftex]{graphicx}
\usepackage{algorithm}
\usepackage{algorithmic}
\usepackage{sidecap}
\usepackage{wrapfig}
\usepackage{colortbl}
\usepackage{tikz}
\usepackage{listings}
\usepackage{xcolor}
\usepackage{makecell}
\usepackage{titletoc}
\usepackage{amsmath}
\usepackage{adjustbox}

\newcount\COMMENTs  
\usepackage{color}
\definecolor{darkgreen}{rgb}{0,0.5,0}
\definecolor{purple}{rgb}{1,0,1}
\newcommand{\comm}[2]{\ifnum\COMMENTs=1\textcolor{#1}{#2}\fi}

\newcommand{\xhdr}[1]{{\noindent\bfseries #1}.}

\definecolor{dkred}{rgb}{0.5,0,0}
\definecolor{dkgreen}{rgb}{0,0.6,0}
\definecolor{gray}{rgb}{0.5,0.5,0.5}
\definecolor{mauve}{rgb}{0.58,0,0.82}

\newcommand{\ie}{\textit{i.e.}}
\newcommand{\eg}{\textit{e.g.}}

\definecolor{customgray}{rgb}{0.3,0.3,0.3}
\definecolor{customgreen}{RGB}{140,211,89}

\usepackage{enumitem}   
\usepackage{amsthm}

\ifx\proof\undefined
\newenvironment{proof}{\par\noindent{\bf Proof\ }}{\hfill\BlackBox\\[2mm]}
\fi
\theoremstyle{plain}
\ifx\theorem\undefined
\newtheorem{theorem}{Theorem}

\theoremstyle{definition}

\fi

\DeclareSymbolFont{extraup}{U}{zavm}{m}{n}
\DeclareMathSymbol{\varheart}{\mathalpha}{extraup}{86}
\DeclareMathSymbol{\vardiamond}{\mathalpha}{extraup}{87}

\title{PyTorch Frame: A Modular Framework for Multi-Modal Tabular Learning}

\author{%
Weihua Hu$^{1}$, Yiwen Yuan$^{1}$, Zecheng Zhang$^{1}$, Akihiro Nitta$^{1}$, Kaidi Cao$^{1}$, \\ 
\bf Vid Kocijan$^{1}$, Jinu Sunil$^{1}$, Jure Leskovec$^{1,2}$, Matthias Fey$^{1}$ \\ \\
$^{1}$Kumo AI, $^{2}$Stanford University
}

\begin{document}

\maketitle

\begin{abstract}
We present PyTorch Frame, a PyTorch-based framework for deep learning over multi-modal tabular data. 
PyTorch Frame makes tabular deep learning easy by providing a PyTorch-based data structure to handle complex tabular data, introducing a model abstraction to enable modular implementation of tabular models, and allowing external foundation models to be incorporated to handle complex columns (\eg, LLMs for text columns).
We demonstrate the usefulness of PyTorch Frame by implementing diverse tabular models in a modular way, successfully applying these models to complex multi-modal tabular data, and integrating our framework with PyTorch Geometric, a PyTorch library for Graph Neural Networks(GNNs), to perform end-to-end learning over relational databases.
\end{abstract}

\section{Introduction}
\label{sec:intro}
\begin{wrapfigure}{r}{0.44\textwidth} 
\vspace{-0.3cm}
\centering
\includegraphics[width=\linewidth]{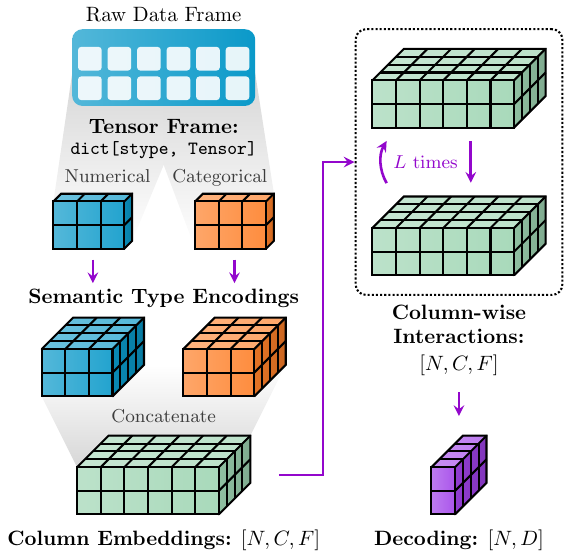}
\caption{\textbf{Overview of PyTorch Frame's architecture}, consisting of a \textbf{(1)} Tensor Frame materialization stage, \textbf{(2)} semantic type-wise model encodings, \textbf{(3)} column-wise interaction blocks, and a final \textbf{(4)} readout decoder head.}
\label{fig:image1}
\end{wrapfigure}

Deep learning has revolutionized many application domains, such as computer vision~\citep{he2016deep}, natural language processing~\citep{brown2020language}, audio processing~\citep{oord2016wavenet}, and graphs~\citep{kipf2016semi}. Yet, one critical domain that has yet to see big success is the \emph{tabular domain}---a powerful and ubiquitous representation of data via heterogeneous columns. In the tabular domain, many existing studies~\citep{shwartz2022tabular,grinsztajn2022tree} have reported that Gradient-Boosted Decision Trees (GBDT)~\citep{chen2016xgboost} is still a dominant paradigm.


However, GBDT has notable limitations. First, GBDT models are primarily focused on numerical and categorical features and cannot effectively handle raw multi-modal features, such as texts, sequences, images, and embeddings.
Second, their end-to-end integration with downstream deep learning models, such as Graph Neural Networks (GNNs), is highly non-trivial since GBDT models are neither differentiable nor producing embeddings~\citep{ivanov2021boost}.
As such, GBDT falls short on complex applications, such as prediction over modern relational databases~\cite{fey2023relational}.

Tabular deep learning is a promising paradigm to resolve the challenges. In fact, the community has come up with many deep tabular models in an attempt to outperform GBDT~\citep{huang2020tabtransformer,gorishniy2021revisiting,gorishniy2022embeddings,gorishniytabr,chen2023excelformer,arik2021,somepalli2021saint,zhu2023xtab,popov2019neural,abutbul2020dnf,chen2023trompt}.
While significant progress has been made, these models have only been evaluated on conventional numerical/categorical features. 
What is missing is a systematic exploration of model architectures and their capabilities in handling complex columns with general multi-modal data.

Here we introduce \emph{PyTorch Frame}, a new PyTorch-based framework for tabular deep learning. Our goal is to facilitate research in tabular deep learning and realize its full potential.
First, realizing the limited expressiveness of vanilla PyTorch to hold multi-modal data, we introduce \emph{Tensor Frame}, an expressive Tensor-based data structure to handle arbitrary complex columns in an efficient way.
Second, we introduce a general framework for learning on tabular data that abstracts the commonalities between the most promising existing deep learning models for tabular data. Our framework is illustrated in Figure~\ref{fig:image1} and shares a similar spirit to the message passing framework~\citep{gilmer2017neural} that has propelled the field of graph learning.
Given that many strong tabular models follow our general framework, we believe the community can further advance modeling with it more easily.

Under our framework, it is easy to incorporate external foundation models to handle complex multi-modal columns. They can be used to either generate embeddings or be finetuned end-to-end with deep tabular models.
Moreover, models implemented with our framework can be easily integrated with other PyTorch models. For instance, by integrating with GNNs from PyTorch Geometric~\citep{fey2019fast}, we can achieve deep learning over relational databases~\citep{fey2023relational}.
Finally, we demonstrate the usefulness of our framework by showing promising results on complex tabular data (\ie~multi-modal columns, multiple tables), in addition to conventional numerical/categorical datasets.

\section{Related Work}
\label{sec:related}
Our framework follows the modular encoder-combiner-decoder framework~\citep{molino2019ludwig}, while being explicit about modeling multi-layer column interactions notable in modern deep tabular models~\citep{chen2023excelformer,gorishniy2021revisiting,chen2023trompt,huang2020tabtransformer}.
Our framework is also related to PyTorch Tabular~\citep{joseph2021pytorch}, an open-source tabular learning framework built on top PyTorch. While PyTorch Tabular has primarily focused on supporting existing tabular models, our PyTorch Frame offers enhanced flexibility for exploring and building novel tabular learning approaches while still providing access to established models. PyTorch Frame further distinguishes itself through support for a wider variety of column modalities and streamlined integration with LLMs.

\section{PyTorch Frame}
\label{sec:frame}
PyTorch Frame \footnote{\url{https://github.com/pyg-team/pytorch-frame}}
provides a unified framework for efficient deep learning over tabular data $\mathbf{T} = [ ( v_1, \ldots, v_C ) ]^N_{n=1}$, which holds data across $C$ columns for every of its $N$ rows.
We denote $T[i, j]$ as the raw value of column $j$ in row $i$.
We also use standard NumPy notations~\citep{harris2020array}, such as $T[:, j]$, $T[i, :]$, $T[[i_1, ..., i_k], :]$, and $T[:, [j_1, ..., j_k]]$.

\xhdr{Semantic Type} Modern tabular data is complex, consisting of a variety of multi-modal columns. To effectively handle such data, PyTorch Frame introduces a \emph{semantic type} that specifies the ``modality'' of each column.
A variety of semantic types are supported to handle diverse columns, such as:

\begin{itemize}[leftmargin=0.6cm,itemsep=0pt, parsep=0pt]
    \item \texttt{numerical} type can be used to handle numerical values, such as price and age columns.
    \item \texttt{categorical} type can be used to handle categorical values, such as gender and educational-level columns.
    \item \texttt{multicategorical} type can be used to handle multi-hot categories, such as a movie genres columns.
    \item \texttt{timestamp} type can be used to handle time columns, such as columns storing the date of events.
    \item Both \texttt{text\_embedding} and \texttt{text\_tokenized} types can be used to handle text data, such as columns storing product descriptions.
    The former pre-encode text into embedding vectors, while the latter enables fine-tuning text model parameters.
    \item \texttt{embedding} type can be used to handle columns storing embedding data, such as pre-computed image embedding vectors.
\end{itemize}

Given tabular data $\mathbf{T}$, PyTorch Frame assumes a semantic type being specified for each of $C$ columns. It can be either inferred based on some heuristics or manually specified by users.
We let $\phi(s)$ denote the mapping from a semantic type $s \in \mathcal{S}$ to the list of column indices specified as $s$.

As shown in Figure \ref{fig:image1}, PyTorch Frame learns representation vectors of $\mathbf{T}$ in the following four stages:\footnote{In practice, we consider a mini-batch of rows $\mathbf{T}[[i_1,...,i_k], :]$, but its extension is straightforward.}

\begin{enumerate}[leftmargin=0.6cm,itemsep=0.5pt, parsep=0pt]
    \item \textbf{Materialization} groups column data according to their semantic type and converts the grouped raw values $\mathbf{T}[:, \phi(s)]$ into a tensor-friendly data $\mathbf{F}_s$ of shape $[N, |\phi(s)|, \ast]$, where the last dimension $\ast$ depends on the specific semantic type $s$.
    We refer the dictionary $\{s: \mathbf{F}_s \}_{s \in \mathcal{S}}$ as a \emph{Tensor Frame} representation of $\mathbf{T}$.
    \item \textbf{Encoding} independently embeds each column value into a $F$-dimensional vector. Specifically, for each semantic type $s$, it embeds input tensor data $\mathbf{F}_s$ of shape $[N, |\phi(s)|]$ into embedding $\mathbf{X}_s$ of shape $[N, |\phi(s)|, F]$. Then, it concatenates  $\{\mathbf{X}_s\}_{s\in \mathcal{S}}$ to obtain the column embedding vector $\mathbf{X}$ of shape $[N, C, F]$.
    \item \textbf{Column-wise Interaction} performs multiple layers of column-wise message passing to enrich each column's representation by the knowledge of other columns. For each layer $\ell = 0,..., L-1$, we update the embedding of column $j$ as follows for each row $i$:
\begin{equation}
\label{eq:column_interaction}
\mathbf{X}^{(\ell + 1)}[i, j, :] \leftarrow f_{\theta} \left(\mathbf{X}^{(\ell)}[i, j, :], \{ \mathbf{X}^{(\ell)}[i, c, :] \}_{1 \leq c \leq C } \right),
\end{equation}
where $\mathbf{X}^{(0)} \leftarrow \mathbf{X}$.
The last-layer column embedding $\mathbf{X}^{(L)}[i,:,:]$ for each row $i$ captures high-order interactions among columns within the row.
    \item \textbf{Decoding} summarizes the last-layer column embeddings $\mathbf{X}^{(L)}[i,:,:]$ to obtain row embeddings $\mathbf{Z}[i,:] = g_{\theta}(\mathbf{X}^{(L)}[i,:,:])$ of shape $[D, ]$, where $D$ is the output dimensionality. The output row embedding $\mathbf{Z}$ can be either sent directly to a prediction head for row-wise prediction or used as input to downstream deep learning models, such as GNNs.
\end{enumerate}

Usually, the materialization is performed as a pre-processing step, and the subsequent three stages have parameters to be learned during training. In what follows, we describe each step in more detail.

\subsection{Materialization}
Data materialization takes care of converting the raw input data in $\mathbf{T}$ into a \emph{Tensor Frame}, a tensor-friendly format that can be efficiently processed in a deep learning pipeline.

The key step is to transform $\mathbf{T}[:, \phi(s)]$ with $N$ rows and $|\phi(s)|$ columns into a tensor data $\mathbf{F}_s$ of shape $[N, |\phi(s)|, *]$, for each semantic type $s \in \mathcal{S}$.
Below we show examples for some representative semantic types.

\xhdr{\texttt{numerical}} $\mathbf{T}[:, \phi(s)]$ is already in numerical form, so it can be directly transformed into a standard floating tensor $\mathbf{F}_s$ of shape $[N, |\phi(s)|]$. We model missing values as \texttt{NaN}.

\xhdr{\texttt{categorical}} $\mathbf{T}[:, \phi(s)]$ usually consists of strings, \eg, ``male'', ``female'', ``non-binary'' in the case of a gender column. For each column, we map elements into an non-negative contiguous indices, \eg, ``male''$\mapsto 0$, ``female''$\mapsto 1$, ``non-binary''$\mapsto 2$. Applying this fixed mapping, we transform $\mathbf{T}[:, \phi(s)]$ into a standard integer tensor $\mathbf{F}_s$ of shape $[N, |\phi(s)|]$. We model missing values as $-1$.

\begin{figure}[t]
\centering
\includegraphics[width=0.6\textwidth]{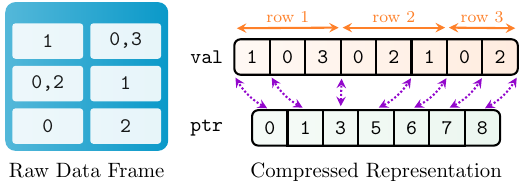}
\caption{\textbf{\texttt{MultiNestedTensor} based on compressed ragged tensor layout.} Our ragged layout describe tensors of shape $[N, C, \cdot]$, where the size of the last dimension can vary across both rows and columns.
Internally, data is stored in an efficient compressed format \texttt{(val, ptr)}, where \texttt{val} holds data in a flattened vector and \texttt{ptr} holds cumulated offsets of rows and columns. $T[i,j]$ can be accessed via $\texttt{val[ptr[C\,*\,i\,+\,j]:ptr[C\,*\,i\,+\,j\,+\,1]]}$, which allows for efficient slicing and indexing along the row dimension.}
\label{fig:multi_nested}
\vspace{-0.3cm}
\end{figure}

\xhdr{\texttt{multicategorical}} Each cell in $\mathbf{T}[:, \phi(s)]$ consists of a list of multiple categories, \eg, [``comedy'', ``romance'', ``drama''] for a movie genre column. We can similarly map each category into an integer index \eg, ``comedy''$\mapsto 0$, ``romance''$\mapsto 1$, ``drama''$\mapsto 2$ so that each cell in $\mathbf{T}[:, \phi(s)]$ can be mapped to a list of integers, \eg, [0, 1, 2] in the case above. The challenge in the implementation lies in the varying sizes of the lists for different cells.
To handle such data, PyTorch Frame supports its own tensor format called \texttt{MultiNestedTensor} based on a \emph{ragged tensor layout} as illustrated in Figure~\ref{fig:multi_nested}. We use it to transform $\mathbf{T}[:, \phi(s)]$ into \texttt{MultiNestedTensor} $\mathbf{F}_s$ of shape $[N, |\phi(s)|, \cdot]$.

\xhdr{\texttt{text\_tokenized}} Each cell of $\mathbf{T}[:, \phi(s)]$ consists of a piece of text, which can be tokenized into a list of integers of varying length. Hence, we can transform $\mathbf{T}[:, \phi(s)]$ into \texttt{MultiNestedTensor} $\mathbf{F}_s$ of shape $[N, |\phi(s)|, \cdot]$, similar to \texttt{multicategorical}. 

\xhdr{\texttt{text\_embedded}} Similar to \texttt{text\_tokenized}, each cell of $\mathbf{T}[:, \phi(s)]$ consists of a piece of text. Different from \texttt{text\_tokenized}, we use external text embedding models~\citep{neelakantan2022text,reimers-2019-sentence-bert} to pre-compute text vectors. Concretely, each single column $\mathbf{T}[:, j], j\in \phi(s)$ is pre-encoded by a column-specific text model into an embedding tensor of shape $[N, 1, D_j]$, where $D_j$ can be different for different $j$. 
To handle multiple text columns simultaneously, PyTorch Frame introduces its own \texttt{MultiEmbeddingTensor} layout, where each column of $\mathbf{T}[:, \phi(s)]$ is pre-embedded into $D_j$-dimensional vectors, which are stacked to produce \texttt{MultiEmbeddingTensor} $\mathbf{F}_s$ of shape $[N, |\phi(s)|, \mathbf{D}]$.
We also support \texttt{image\_embedded} in the similar way, whereas each cell contains the path of the image data.

\xhdr{\texttt{embedding}}
A table may contain pre-computed embeddings, such as those created by other teams~\citep{hu2022learning}. Specifically, $\mathbf{T}[:, j], j\in \phi(s)$ directly stores $D_j$-dimensional embeddings. Similar to \texttt{text\_embedded}, we can transform $\mathbf{T}[:, \phi(s)]$ into a \texttt{MultiEmbeddingTensor} $\mathbf{F}_s$ of shape $[N, |\phi(s)|, \mathbf{D}]$.

In summary, PyTorch Frame uses specialized tensor-based data structures to efficiently handle complex tabular data with different semantic types. 
In addition, the materialization stage computes basic statistics for each column, such as mean and standard deviation for \texttt{numerical} columns, or the count of category elements for \texttt{categorical} and \texttt{multicategorical} columns.
These statistics are stored and supplied to the subsequent encoding stage to normalize data or impute missing values.

\subsection{Encoding}
PyTorch Frame encoders receive a Tensor Frame with $N$ rows as input\footnote{In mini-batch training, they receive a Tensor Frame of $B \leq N$ rows.} and map their columns into a shared embedding space $\mathbf{X}$ of shape $[N, C, F]$.
All columns within the same semantic type are embedded in parallel, ensuring maximum throughput. More concretely, for each semantic type $s$, its tensor data $\mathbf{F}_s$ of shape $[B, |\phi(s)|, *]$ is embedded into $\mathbf{X}_s$ of shape $[B, |\phi(s)|, F]$, where $F$ is the dimensionality of column embeddings. Then, $\{\mathbf{X}_s\}_{s\in \mathcal{S}}$ are concatenated to produce the final shared embedding $\mathbf{X}$ of shape $[N, C, F]$.
In mapping to $\mathbf{X}_s$, encoders perform feature normalization and column embeddings, as detailed below.

\xhdr{Feature normalization}
The tensor data $\mathbf{F}_s$ can contain missing and values with arbitrary scales, making them not suitable as input to machine learning models. To resolve these issues, encoders first normalize the features based on the statistics calculated from the materialization stage.
As an example, for \texttt{numerical} type, one can impute missing values with the mean value. Then, each column can be normalized to have zero-mean and unit-variance.
The feature normalization is performed at the encoding stage (instead of at the materialization stage), which allows our users to test various imputation/normalization strategies without the need to re-materialize the data.

\xhdr{Column Embeddings}
After feature normalization, the encoders embed $\mathbf{F}_{s}$, representing $|\phi(s)|$ columns, into $F$-dimensional column embeddings $\mathbf{X}_j$ of shape $[N, |\phi(s)|, F]$. 
Different modeling choices are possible.
For example, \texttt{numerical} columns can either be transformed either via a linear layer or can be first converted into piecewise linear or periodic representations~\cite{gorishniy2022embeddings} before a linear layer.
For \texttt{categorical} columns, one can transform them via shallow embeddings learnable for each category.
The \texttt{text\_tokenized} columns can be transformed into embeddings via language models that take the sequences of tokens as input.

\subsection{Column-wise Interaction}
Given the column embeddings $\mathbf{X}$, where all columns are embedded in a shared $F$-dimensional embedding space, we proceed to model the interactions between columns in the embedding space.
Specifically, an embedding of each column are iteratively updated based on those of the other columns, as shown in Eq.~\ref{eq:column_interaction}. After $L$ iterations, we obtain $\mathbf{X}^{(L)}$ of shape $[N, C, F]$, capturing higher-order interactions among column values within each row.

Many existing works can be cast under this framework. For example, \citet{gorishniy2021revisiting} applied a permutation-invariant Transformer~\citep{vaswani2017attention} to model column interactions, while \citet{huang2020tabtransformer} used a Transformer with positional column encoding.
\cite{chen2023excelformer} also followed a Transformer architecture except that it sorts the features by mutual information and used diagonal attention in the Transformer block.
\citet{chen2023trompt} used cross attention between column embeddings and learnable prompt embeddings to model column interactions.

\subsection{Decoding}
Finally, we apply a decoder on $\mathbf{X}^{(L)}$ to obtain $D$-dimensional row-wise embedding $\mathbf{Z}$ of shape $[N, D]$, which can be directly used for prediction over tabular rows or as input to subsequent deep learning models.

The decoder can be, \emph{e.g.}, a weighted sum of column embeddings, where the weights are either uniform or learned attention weights~\citep{chen2023trompt}. \citet{huang2020tabtransformer} modeled the decoder by applying an MLP over the flattened column embeddings of length $C \times F$. \citet{gorishniy2021revisiting} added a ``CLS'' column embedding~\citep{devlin2018bert} in $\mathbf{X}$ and directly read out ``CLS'' column embeddings in $\mathbf{X}^{(L)}$, similar to BERT~\citep{devlin2018bert}.

\subsection{Accommodating Diverse Tabular Models}
While our abstraction framework covers many existing tabular models, not all models fit within our framework. Some models are simple to accommodate, \eg, ResNet~\citep{gorishniy2021revisiting} does not have the column-wise interaction stage, so we can simply omit the stage. Other models are harder to accommodate. For example, TabNet~\citep{arik2021} operates on 2-dimensional tensors (instead of the 3-dimensional tensor layout of $\mathbf{X}$) and applies a series of attention-based transformation over it.
Notably, those models can still be implemented by taking Tensor Frame as input; PyTorch Frame supports those models by directly implementing the model architecture without following the modular framework.

\section{Integration}

\xhdr{Integration with Foundational Models}
For modeling complex columns like text and images, it is best to incorporate external large pre-trained foundation models.
PyTorch Frame supports seamless integration with external models via semantic types like \texttt{text\_embedded}, \texttt{text\_tokenized}, and \texttt{image\_embedded}.

For example, for \texttt{text\_embedded}, users only need to specify embedding models to map a list of texts into embedding tensors, which can be achieved via the OpenAI embedding API\footnote{\url{https://platform.openai.com/docs/guides/embeddings}} or any sentence transformer~\citep{reimers-2019-sentence-bert}. Then, at the materialization stage, PyTorch Frame automatically applies the embedding models to generate a Tensor Frame with text embeddings. 
Note that the materialization can be expensive since it uses LLMs to embed all text columns. To avoid repeated materialization, PyTorch Frame \emph{caches} the materialized data. The cached Tensor Frame can be reused in  subsequent runs, avoiding expensive re-materialization. 

\begin{figure*}[t]
\centering
\includegraphics[width=1.0\textwidth]{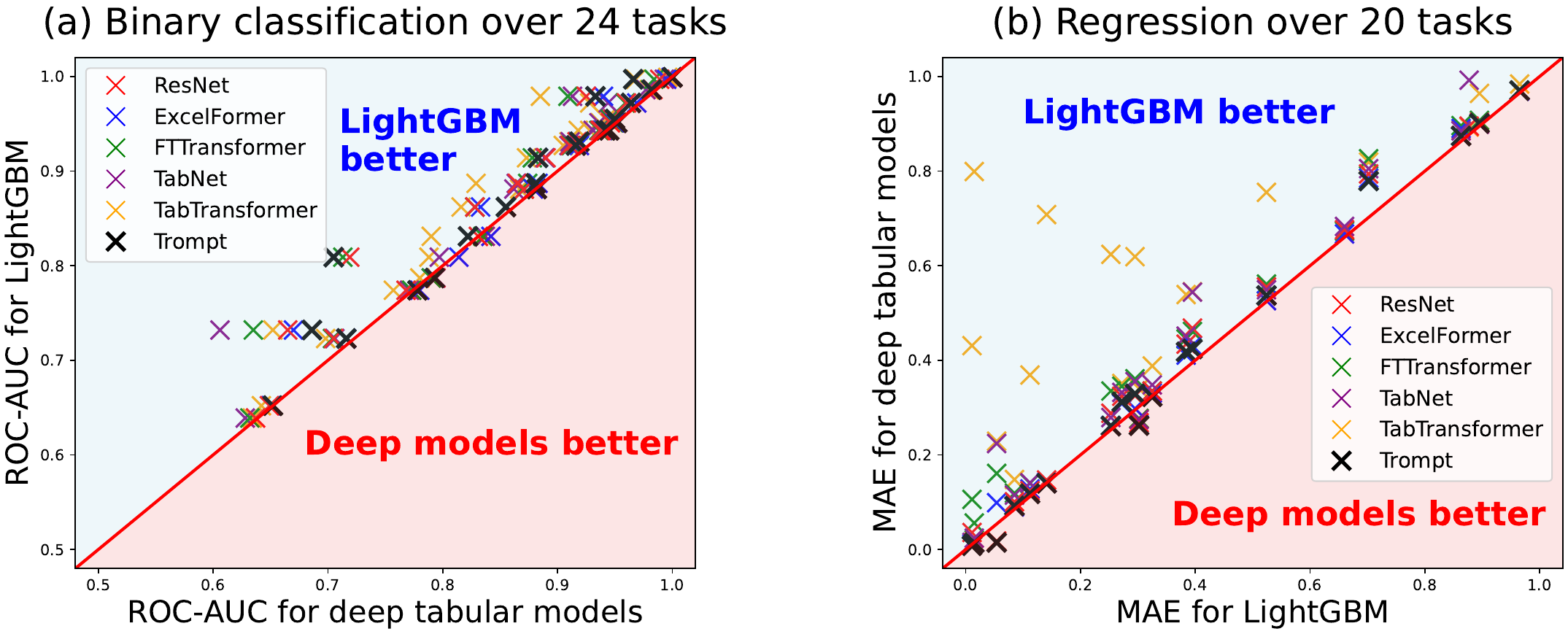}
\caption{\textbf{Scatter plot comparison between deep tabular models and LightGBM across datasets with \emph{only numerical and categorical features}.} Here each ``x'' represents a single predictive task, and its position represents the predictive performance of a deep tabular model compared against LightGBM. When ``x'' lies above (resp.~below) the diagonal line, it means the LightGBM outperforms (resp.~underperforms) the corresponding deep tabular model on the respective task. Overall, LightGBM is still dominating the existing deep tabular models on the conventional numerical/categorical datasets, although the recent Trompt model~\citep{chen2023trompt} is getting close.}
\label{fig:deep_xgboost_comp}
\vspace{-0.3cm}
\end{figure*}

\xhdr{Integration with PyTorch Geometric}
We have so far discussed single-table tabular learning, but many practical applications involve data stored in a \emph{relational format}~\citep{codd1970relational}, where tabular data is connected with each other via primary-foreign key relationships. Combining tabular deep learning with Graph Neural Networks (GNNs) has proven to be promising to handle such relational datasets~\cite{fey2023relational}. PyTorch Frame integrates natively with PyTorch Geometric (PyG)~\cite{fey2019fast}, a popular PyTorch library for GNNs. PyTorch Frame enhances PyG by learning embedding vectors of nodes and edges with complex multi-modal features. The node and edge embeddings are subsequently fed as input to GNNs by PyG. 
Crucially, tabular deep learning models by PyTorch Frame and GNNs by PyG can be jointly trained to optimize for downstream task performance.

\section{Experimental Study}
\label{sec:experiments}
Here we demonstrate the usefulness of PyTorch Frame in handling conventional single-table data as well as more complex tabular data with text columns and relational structure.

\subsection{Handling single-table data}
\label{subsec:num_cat_exp}

First, we focus on the traditional tabular machine learning setting with only numerical and categorical columns. We collected datasets from diverse resources~\citep{grinsztajn2022tree,gorishniy2021revisiting,blake1998uci}, totalling 23 tasks for binary classification and 19 tasks for regression.
Following \citet{hu2020open}, we make all these datasets and and their data split available through PyTorch Frame package so that it is easy to compare models in a standardized manner.

Using PyTorch Frame, we implemented six deep tabular models: ResNet~\citep{gorishniy2021revisiting}, ExcelFormer~\citep{chen2023excelformer}, FTTransformer~\citep{gorishniy2021revisiting}, TabNet~\citep{arik2021}, TabTransformer~\citep{huang2020tabtransformer}. PyTorch also seamlessly integrated GBDT models, XGBoost~\citep{chen2016xgboost}, CatBoost~\citep{prokhorenkova2018catboost}, and LightGBM~\citep{ke2017lightgbm}, that operate on Tensor Frame.
For each model, we used Optuna to perform a hyper-parameter search with 20 trials~\citep{akiba2019optuna}. 

Figure~\ref{fig:deep_xgboost_comp} shows the comparison between each of the six deep learning models and LightGBM, which we found to perform the best among the GBDT models.
Similar to previous studies~\citep{shwartz2022tabular,grinsztajn2022tree}, we found that deep tabular models are coming close to LightGBM, but not outperforming it. Among the six deep tabular models, we found the Trompt model to give the closest performance to LightGBM, but Trompt is also the most expensive deep tabular model with nearly 100 to 1000 times more training time compared to LightGBM even with GPU. Given the simplicity and efficiency of GBDT models, they may remain a practical choice for conventional tabular learning datasets.

\begin{table}[t]
  \centering
  \caption{
    \textbf{Results on tabular datasets with text columns.} We report binary classification ROCAUC and training time (excluding text pre-encoding). For each dataset, bolded values represent the best in each category, with $^*$ indicating the best overall. Details of text models: 
    ${\spadesuit}$ \texttt{all-roberta-large-v1} (Sentence Transformer)~\citep{reimers-2019-sentence-bert}. 
    ${\clubsuit}$ \texttt{text-embedding-3-large} (OpenAI)~\citep{neelakantan2022text}. 
    ${\vardiamond}$ RoBERTa-large~\citep{liu2019roberta}. 
    ${\varheart}$ best model from \citet{shi2020masked} (RoBERTa-large or ELECTRA~\citep{clark2020electra}). 
    For LightGBM$^{\dagger}$, text embeddings are treated as numerical features.
  }
  \vspace{0.1cm}
  \label{tab:text-bench-res}
  \renewcommand{\arraystretch}{1.0}
  \setlength{\tabcolsep}{2pt}
  \begin{tabular}{llcrcrcr}
    \toprule
    \multicolumn{2}{c}{\textbf{Method}} & \multicolumn{2}{c}{\textbf{fake}} & \multicolumn{2}{c}{\textbf{jigsaw}} & \multicolumn{2}{c}{\textbf{kick}} \\
    \cmidrule(lr){1-2} \cmidrule(lr){3-4} \cmidrule(lr){5-6} \cmidrule(lr){7-8}
    \textbf{Text Model} & \textbf{Tabular Model} & ROC-AUC & Time & ROC-AUC & Time & ROC-AUC & Time \\
    \midrule
    \multirow{4}{*}{\texttt{RoBERTa}$^{\spadesuit}$} 
    & ResNet             & 0.934  & 7.3s    & 0.883  & 36.1s    & 0.753  & 27.8s \\
    & FTTransformer      & 0.936  & 19.7s   & 0.882  & 100.8s   & 0.747  & 77.4s \\
    & Trompt             & \textbf{0.958}  & 18.8s   & \textbf{0.885}  & 480.4s   & 0.756  & 581.9s \\
    & LightGBM$^{\dagger}$ & 0.954 & 15.5s   & 0.865  & 571.1s   & \textbf{0.767} & 1931.9s \\
    \midrule
    \multirow{4}{*}{\texttt{OpenAI}$^{\clubsuit}$} 
    & ResNet             & 0.923  & 10.4s   & 0.945  & 56.5s    & 0.807  & 107.1s \\
    & FTTransformer      & 0.911  & 23.6s   & 0.945  & 337.4s   & 0.807  & 168.9s \\
    & Trompt             & \textbf{0.976}  & 40.8s   & \textbf{0.947}  & 4285.1s  & \textbf{0.810}$^*$ & 538.0s \\
    & LightGBM$^{\dagger}$ & 0.966 & 131.0s  & 0.926  & 1732.9s  & 0.809  & 1924.3s \\
    \midrule
    \texttt{RoBERTa}$^{\vardiamond}$ & ResNet & \textbf{0.979}$^*$ & 5.5h & \textbf{0.970}$^*$ & $>$1 day & \textbf{0.786} & $>$1 day \\
    & FTTransformer & 0.960 & 5.5h & 0.968 & $>$1 day & 0.775 & $>$1 day \\
    \midrule
    \multicolumn{2}{l}{Best single model~\citep{shi2021benchmarking}$^{\varheart}$} & 0.967 & - & 0.967 & - & 0.794 & - \\
    \bottomrule
  \end{tabular}
\end{table}

Next, we shift our evaluation to more modern tabular datasets that come with text columns and multiple tables.

\subsection{Handling text data}
In this section, we demonstrate the capability of PyTorch Frame in utilizing external text models to achieve strong performance on tabular datasets with text columns.

PyTorch Frame provides two options to handle text columns: \texttt{text\_embedded} and \texttt{text\_tokenized}.
The \texttt{text\_embedded} option pre-encodes text into embedding vectors at the materialization stage, while \texttt{text\_tokenized} option only tokenizes text during materialization, allowing text models to be jointly trained with deep tabular models at training time.

For \texttt{text\_embedded}, we consider two kinds of text embedding models: The \texttt{all-roberta-large-v1} model from the Sentence Transformer~\citep{liu2019roberta, reimers-2019-sentence-bert} and the more recent OpenAI embedding model, \texttt{text-embedding-3-large}, available through API~\citep{neelakantan2022text}.\footnote{Note that we need to be aware that the OpenAI embedding model may be trained on the experimented tabular data.} For \texttt{text\_tokenized}, we used the original RoBERTa-large model~\citep{liu2019roberta}, to align with the setting in \citet{shi2021benchmarking}.
We trained strong deep tabular models\footnote{We did not include the Trompt model in our \texttt{text\_tokenized} experiment since the model architecture requires applying the text model in each layer, which is very GPU memory intensive.} and LightGBM to make the final label prediction. The hyper-parameters of LightGBM are tuned with Optuna~\citep{akiba2019optuna} with 3 trials, while those of deep tabular models are tuned manually.

The results are shown in Table~\ref{tab:text-bench-res}.
Overall, we find that the best results from PyTorch Frame significantly improve over the best single-model results from \citet{shi2021benchmarking}, demonstrating the promise of PyTorch Frame in handling tabular data with text columns.

Comparing among models with \texttt{text\_embedded}, we see clear benefit of using the advanced OpenAI embedding model as opposed to the less advanced \texttt{RoBERTa} model. Moreover, the Trompt model often provides the best performance among the tabular models, collaborating our finding in Section~\ref{subsec:num_cat_exp}.

Comparing between \texttt{text\_embedded} and \texttt{text\_tokenized} options with the same base text model (\ie, RoBERTa), we see that \texttt{text\_tokenized} gives a substantially better predictive performance. 
This is expected since \texttt{text\_tokenized} allows the text model to be specifically fine-tuned on predictive tasks of interest.
However, \texttt{text\_tokenized} is orders-of-magnitude slower than \texttt{text\_embedded} due to the expensive fine-tuning of text models at training time. Nonetheless, by using the more advanced OpenAI embeddings, \texttt{text\_embedded} gives significantly better performance that is comparable to that of the \texttt{text\_tokenized} option, while being much faster than \texttt{text\_tokenized} in terms of training time. With the faster and cheaper text embedding API available, the \texttt{text\_embedded} option becomes a promising choice to achieve good performance on tabular datasets with text columns.

\subsection{Handling relational data}
Finally, we show the benefit of tabular deep learning by integrating PyTorch Frame models with PyG~\citep{fey2019fast}
to make predictions over relational databases. 

We consider \texttt{rel-stackex}, a Stack Exchange dataset from ~\citet{fey2023relational}. It consists of 7 tables that store users, posts, comments, votes, post links, badge records, and post history records. Within the dataset, two practically relevant prediction tasks are defined. The \texttt{rel-stackex-engage} aims to predict if the user will make any contribution, defined as vote, comment, or post, in the next 2 years. 
The \texttt{rel-stackex-votes} task aims to predict the popularity of a question post in the next 2 years, where the popularity is defined as the number of upvotes the post will receive. \texttt{rel-stackex-engage} is a binary classification task, while \texttt{rel-stackex-votes} is a regression task.

Following the relational deep learning approach~\citep{fey2019fast}, we use deep tabular models to encode table rows into node embeddings, which are then fed into GNNs to update the embeddings based on primary-foreign key relations. Crucially, the deep tabular models and GNNs are \emph{jointly trained} to optimize for the task performance.
As a specific instantiation, we adopted ResNet from PyTorch Frame for row encoding and heterogeneous GraphSAGE from PyG for updating node embeddings. We compare our model against a LightGBM that is trained on a single table data. As we see in Table~\ref{tab:rel-bench-res}, the relational deep learning approach enabled by the combination of PyTorch Frame and PyG provides superior performance compared to LightGBM that can be only trained on single-table data.

\begin{table}[t]
  \centering
  \caption{\textbf{Results on multi-tabular relational datasets}.}
  \vspace{0.1cm}
  \label{tab:rel-bench-res}
  \renewcommand{\arraystretch}{0.9}
  \begin{tabular}{lccc}
    \toprule
     \mr{3}{Method} & \mc{2}{c}{\texttt{rel-stackex-}} \\
                    & \texttt{engage} & \texttt{votes} \\
                    & ROCAUC $\uparrow$ & MAE $\downarrow$ \\
    \midrule
    LightGBM & 0.618 & 0.422 \\
    \midrule
    PyG-HeteroSAGE+ & \mr{2}{\textbf{0.854}} & \mr{2}{\textbf{0.373}}\\
    PyTorch Frame-ResNet &   \\
    \bottomrule
  \end{tabular}
\end{table}

\section{Conclusions}
\label{sec:conclusion}
We presented PyTorch Frame to facilitate deep learning research on tabular data. We introduced Tensor Frame, a new tensor-based data structure to efficiently handle multi-modal tabular data. Then, we built a general model abstraction on top of Tensor Frame and implemented state-of-the-art deep tabular models under the modular framework.
We empirically demonstrate the usefulness of PyTorch Frame on modern tabular learning settings involving text columns and multiple tables.
Overall, we hope PyTorch Frame helps pushing tabular deep learning to enable accurate prediction over complex multi-modal tabular data.

\bibliography{reference}
\bibliographystyle{icml2024}

\newpage

\appendix
\onecolumn

\end{document}